\documentclass{article}

\PassOptionsToPackage{numbers, sort&compress}{natbib}
\usepackage[preprint]{neurips_2026}

\usepackage[utf8]{inputenc} 
\usepackage[T1]{fontenc}    
\usepackage{hyperref}       
\usepackage{url}            
\usepackage{booktabs}       
\usepackage{amsfonts}       
\usepackage{nicefrac}       
\usepackage{microtype}      
\usepackage{xcolor}         
\usepackage{graphicx}
\usepackage{tabularx}
\usepackage{amsmath}

\newif\ifshowbswrsignaltable
\showbswrsignaltablefalse

\title{On Sleep–Wake Disorder Identification from Polysomnographic Representations}

\author{Alvise Dei Rossi \\
  Faculty of Informatics, Università della Svizzera Italiana \\
  MeDiTech, SUPSI \\
  Lugano, Switzerland \\
  \texttt{alvise.dei.rossi@usi.ch} \\
  \AND
  Luigi Fiorillo \\
  MeDiTech, SUPSI \\
  Ente Ospedaliero Cantonale (EOC) \\
  Lugano, Switzerland \\
  \texttt{luigi.fiorillo@supsi.ch} \\
  \And
  Giuliana Monachino \\
  MeDiTech, SUPSI \\
  Lugano, Switzerland \\
  \texttt{giuliana.monachino@supsi.ch} \\
  \AND
  Alessia Calzoni \\
  Department of Information Engineering, University of Brescia \\
  Brescia, Italy \\
  \texttt{alessia.calzoni@unibs.it} \\
  \AND
  Julia van der Meer \& Markus H. Schmidt \& Claudio L.A. Bassetti \\
  Sleep Wake Epilepsy Center, Department of Neurology, Inselspital \\
  University of Bern \\
  Bern, Switzerland \\
  \texttt{\{julia.vandermeer, markus.schmidt, claudio.bassetti\}@insel.ch} \\
  \AND
  Silvia Santini \\
  Faculty of Informatics, Università della Svizzera Italiana \\
  Lugano, Switzerland \\
  \texttt{silvia.santini@usi.ch} \\
  \And
  Francesca Faraci \\
  MeDiTech, SUPSI \\
  Lugano, Switzerland \\
  \texttt{francesca.faraci@supsi.ch}
}

\begin{document}

\maketitle

\begin{abstract}
Self-supervised polysomnography foundation models (PSG FMs) aim to capture clinically useful information beyond sleep staging, yet are rarely compared with sleep-staging encoders pretrained at comparable scale and modality coverage. Using a common whole-night, missingness-aware state-space downstream modeling framework, we compared frozen representations from prominent large-scale, multi-cohort sleep-stage-supervised encoders and self-supervised PSG FMs for multilabel identification of six sleep-wake and circadian disorder groups in 7,906 PSGs from an unseen registry. Although PSG FMs outperformed an EEG-only, single-cohort supervised baseline, this ordering reversed when supervised pretraining approached comparable scale and modality coverage. Transfer learning from neural sleep scorers achieved higher overall performance while exhibiting greater downstream data efficiency. These findings underscore the need for scale- and modality-matched pretrained supervised baselines when assessing PSG FMs' downstream potential.
\end{abstract}

\section{Introduction}
Polysomnography (PSG), the reference standard for comprehensive overnight assessment of sleep physiology, is a central diagnostic tool across multiple sleep-wake-circadian disorders (SWCDs) and synchronously records brain (EEG), ocular (EOG), muscular (EMG), cardiac (ECG), and respiratory activity (RESP) \citep{Kushida2005,Ibanez2018}. From a computational perspective, PSG represents both a multimodal and a long-sequence modeling problem: diagnostically relevant information ranges from transient waveform events to interactions among physiological signals and changes in sleep architecture unfolding over hours. Clinical interpretation renders this complexity actionable by assigning each 30-second segment to wake, N1, N2, N3, or REM sleep stages \citep{Berry2017}, annotating physiological events, and deriving night-level summary measures. These abstractions provide readily interpretable descriptions of sleep, but continuous, multimodal PSG signals may contain complementary information associated with SWCD phenotypes that discrete scores or aggregated metrics do not fully capture. Accordingly, we evaluate representations from currently available pretrained PSG encoders for SWCD identification, considering differences in pretraining strategy, scale, and physiological coverage.

Early computational approaches to PSG analysis relied predominantly on handcrafted time-, frequency-, and nonlinear-domain features combined with conventional classifiers \citep{Fiorillo2019, Vallat2021}. Deep learning progressively shifted the field toward end-to-end representation learning  from raw signals or time--frequency inputs, initially driven largely by automatic sleep staging, i.e., the determination of the sequence of sleep stages across the night \citep{Phan2022Review}. Together with the growing availability of large annotated PSG collections \citep{Zhang2024}, this transition enabled sleep-staging models to learn from heterogeneous populations and recording configurations and to generalize to unseen cohorts \citep{Perslev2021,DeiRossi2025,Naps2026}.

Recent research on PSG representations has increasingly centered on models described as foundation models (FMs): encoders pretrained at scale, typically through self-supervised objectives, and intended for reuse across multiple downstream tasks \cite{Helmy2026}. Motivated by the cost and variability of expert annotations and by the aim of learning physiological structure beyond sleep staging, recent approaches use cross-modal contrastive learning, self-distillation, and autoregressive next-token prediction \citep{Thapa2026,Shuai2026,Carter2026}. Their reported transfer across sleep staging, physiological-event detection, and prediction of a wide array of disorders has established self-supervised pretraining as a prominent direction for PSG representation learning.

However, the role of self-supervision itself for downstream transfer should be clarified; although self-supervision often enables scale when task-relevant annotations are scarce, PSG is unusual in this respect: research-accessible repositories, like the National Sleep Research Resource (NSRR) \citep{Zhang2024}, contain abundant human-scored sleep-stage annotations for supervised pretraining across multiple cohorts. This places large-scale supervised efforts \citep{DeiRossi2025} within the same broad range as most recent PSG FMs in terms of data scale and diversity, with the exception of SleepFM, which was pretrained on roughly twice as much PSG data \citep{Thapa2026}. Moreover, neural sleep scorers can retain clinically relevant information beyond their discrete stage predictions: for instance, their latent representations, probabilistic outputs, and temporal dynamics have supported the identification of narcolepsy, obstructive sleep apnea, and the prediction of future cardiovascular risk \citep{Stephansen2018,Perslev2021,Deng2025,Bechny2025}.

More fundamentally, recent work on the pretraining and scaling of PSG FMs identified robustness to missing channels, channel-invariant feature learning, pretraining scale, and multi-source data diversity as key determinants of downstream performance \citep{Shuai2026}. These properties are not unique to self-supervision and echo findings and design principles already established in automatic sleep staging: vulnerability to missing or changing channel configurations \citep{Phan2022Review}, the benefit of channel-agnostic training across variable sensor derivations \citep{Perslev2021}, and the importance of data scale and cohort diversity for generalization \citep{Olesen2021,DeiRossi2025}.
Nevertheless, transfer learning of latent representations from large, multi-cohort, sleep-stage-supervised encoders has rarely been evaluated alongside self-supervised FMs. 
Available comparisons have instead often used supervised architectures trained end-to-end on downstream labels, or demographic-only baselines, without including a supervised-pretraining condition \citep{Thapa2026,Shuai2026}. 
In other works, supervised models have been trained separately within each target cohort for staging itself \citep{Carter2026}. 
When frozen supervised representations have been considered for broader
downstream tasks, the evaluated checkpoints were pretrained on a single cohort, and the downstream evaluation considered EEG alone \citep{Kontras2026}.
An important recent exception matched the architecture and pretraining corpus of self-supervised and sleep-stage-supervised models, finding stronger participant cross-night fingerprinting and finer-grained within-record latent organization after self-supervised pretraining \citep{Coon2026}. However, its clearest downstream advantages over the supervised counterpart mostly concerned the prediction of demographic and anthropometric traits, leaving uncertain whether this additional information supports consistent gains across broader clinical outcomes. 

Evaluation protocols also differ in temporal handling, including whole-recording mean embeddings \citep{Kontras2026}, prediction from epoch-level outputs \citep{Carter2026}, and averaged predictions from independently processed segments \citep{Coon2026}. These choices compound differences in pretraining scale and diversity, modality coverage, and encoder architecture, making comparisons across released models difficult to interpret. Fully factorial retraining across pretraining objectives and corpora, encoder architectures, modality configurations, and optimization regimes would be required to isolate their individual effects, but is infeasible at the scale of current PSG frameworks.
In the present work we investigate the pragmatic question of how prominent supervised and self-supervised PSG encoders presently available to practitioners compare on the same unseen clinical cohort. We evaluate their representations for multilabel SWCD identification in a large clinical cohort, reducing avoidable downstream differences through a common sequential modeling framework, which leverages each encoder's temporally resolved whole-night output and, where available, its modality structure. Models are evaluated both with and without demographic covariates, and contextualized against baselines leveraging sleep summary metrics from manual and automated sleep staging, while key differences in pretraining scale and modality coverage are summarized. To our knowledge, this is the first large-scale evaluation of pretrained PSG representations for clinically adjudicated, co-occurring SWCDs, providing a clinically grounded assessment of whether current self-supervised sleep FMs offer greater downstream utility than large-scale, multi-cohort, sleep-stage-supervised encoders for SWCD identification.

\begin{figure*}[t!]
    \centering
    \includegraphics[width=.965\textwidth]{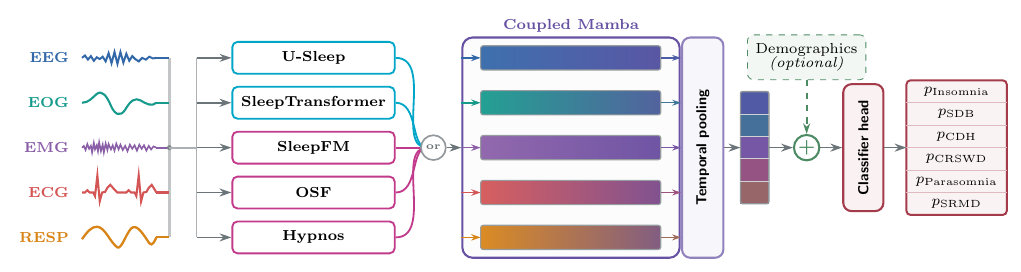}
    \caption{
        Recordings are processed by a pretrained encoder, yielding representations that are jointly modeled across time and streams over the complete night using a mask-aware state-space network. 
    }
    \label{fig:framework}
\end{figure*}

\section{Methods}

\textbf{Dataset \space}
The study cohort comprised 7,906 overnight PSG recordings from 7,195 patients (mean age 48.8 years [SD 18.0]; 33.5\% female) in the Bern Sleep-Wake Registry (BSWR), a clinical registry of patients evaluated during routine sleep medicine practice \citep{vanderMeer2026}. 
Clinical diagnoses were grouped into six categories: insomnia, sleep-disordered breathing (SDB), central disorders of hypersomnolence (CDH), circadian rhythm sleep-wake disorders (CRSWD), parasomnias, and sleep-related movement disorders (SRMD). Recording-level prevalences were 14.3\%, 70.3\%, 10.1\%, 1.2\%, 5.1\%, and 8.3\%, respectively. Labels were non-mutually exclusive: 1,576 recordings (19.9\%) were associated with at least two diagnoses. We therefore formulated SWCD identification as multilabel classification. BSWR did not contribute to the pretraining data of any evaluated encoder and thus constituted an unseen cohort for all models. Detailed cohort and signal characteristics are reported in Appendix~\ref{app:cohort}.

\textbf{Hypnogram-derived baselines \space}
Conventional whole-night sleep-architecture features were derived independently from clinician-provided hypnograms and from automatic hypnograms produced by SOMNUS, a soft-voting ensemble of neural sleep scorers released in SLEEPYLAND \citep{DeiRossi2025}. 
Each feature set comprised total sleep time, wake after sleep onset, sleep-onset and REM latencies, sleep fragmentation, sleep-stage percentages, and 25 stage-transition proportions. 
For each disorder, random forest models were trained using each source alone and in combination with age, gender, BMI, and waist-to-height ratio, together with a covariate-only baseline (Appendix~\ref{apd:rf}).

\textbf{Supervised pretrained encoders \space}
We evaluated U-Sleep \citep{Perslev2021} and SleepTransformer \citep{Phan2022sleeptransformer} checkpoints pretrained under the SLEEPYLAND protocol using approximately 220,000 hours of expert-annotated PSG from 17 NSRR cohorts \citep{DeiRossi2025}. U-Sleep is a convolutional U-Net model that operates on raw physiological signals and produces representations at multiple temporal scales; we retained its pre-classifier representation at one-second resolution. SleepTransformer employs a hierarchical attention-based architecture: an epoch-transformer encodes time-frequency representations within each 30-second epoch, while a sequence-transformer models dependencies across epochs; we retained its pre-classifier output at 30-second resolution. For both architectures, single-channel EEG, EOG, EMG, ECG, and RESP checkpoints were trained across available derivations and applied to eligible channels. We used the EEG and EOG checkpoints released in SLEEPYLAND, while the EMG, ECG, and RESP counterparts were trained under the same protocol, as described in Appendix~\ref{app:supervised_encoders}.

\textbf{Self-supervised pretrained FMs \space}
We selected SleepFM, OSF, and Hypnos as prominent released PSG FMs pretrained on large, multi-cohort collections and representative of distinct self-supervised objectives.
SleepFM organizes signals into four modality groups: brain activity signals (EEG and EOG), cardiac activity, muscular activity, and respiration \citep{Thapa2026}. A shared convolutional encoder maps five-second windows from each channel to 128-dimensional tokens, which a spatial pooling transformer block aggregates within each modality while accommodating variable channel numbers. A temporal transformer contextualizes the resulting sequences over five-minute intervals, and leave-one-out contrastive learning aligns each temporally pooled modality representation with the average of the remaining modalities. We retained the contextualized outputs before temporal pooling, yielding a sequence per modality at five-second resolution.
OSF adapts DINO-style teacher-student self-distillation to multimodal PSG \citep{Oquab2023,Shuai2026}. Channel and temporal masking generate alternative views of the same epoch, encouraging robustness to diverse montages. A vision transformer jointly encodes channel-time patches from each sleep epoch of a fixed 12-channel montage. We retained the CLS embedding before the projection head, yielding one fused representation every 30 seconds.
Hypnos formulates PSG pretraining as autoregressive next-token prediction \citep{Carter2026}. Five tokenizers discretize eight physiological streams into residual-quantization codes at one-second resolution. An RQ-Transformer alternates temporal and cross-modal attention to predict subsequent codes across time and quantization depth, while random modality grouping exposes the model to different combinations of signals. We retained the output before residual-depth prediction, yielding modality-indexed sequences at one-second resolution. Encoder and extraction details are provided in Appendix~\ref{app:self_supervised_encoders}.

\textbf{Shared downstream protocol \space}
All pretrained encoders remained frozen, and their extracted representations were evaluated using separate instances of the same downstream sequence model (Figure \ref{fig:framework}). Representations were supplied in their native organization and temporal resolution, right-padded within each batch, and accompanied by masks identifying valid time points and available streams.

At the finest one-second resolution, a full-night PSG can contain tens of thousands of representation tokens, making the quadratic complexity of full self-attention unattractive for whole-night modeling. State-space models instead propagate information through a fixed-dimensional latent state and scale linearly with sequence length. Mamba introduced input-dependent selective state updates, allowing the model to retain, suppress, or expose information according to the current token while remaining computationally efficient \citep{Gu2023}. These properties motivated its use as our backbone.
To capture interactions among synchronized physiological streams, we used a mask-aware adaptation of Coupled Mamba \citep{Li2024}. Each stream was projected independently to a common width and retained its own residual sequence and Mamba parameterization. Within each block, the input-dependent transition and update terms produced by the available streams were averaged to evolve a shared recurrent state. Each stream then applied its own current transition and update to the preceding shared state, producing a modality-specific output that was written back into its residual token sequence. This enabled information exchange across the night while preserving separate stream outputs; unavailable streams were excluded from shared-state averaging. For OSF, whose encoder already produced one fused representation sequence, the same formulation reduced to a single-stream selective state-space model.
Following whole-night processing, the stream sequences were temporally pooled independently, concatenated, and jointly normalized. Demographic and anthropometric covariates were optionally appended only after temporal pooling. A classification head jointly produced logits for the six diagnostic categories and was optimized using binary cross-entropy.
Implementation details are provided in Appendix~\ref{app:downstream}. For brevity, each downstream model is hereafter named by its representation source (e.g., U-Sleep denotes Coupled Mamba applied to frozen U-Sleep representations).

\textbf{Evaluation \space}
All methods were evaluated using the same fixed five-fold cross-validation, with all recordings from each patient assigned to the same fold. Per-disorder AUROC was computed from pooled out-of-fold predictions, with macro-AUROC as a secondary summary.
Targeted experiments examined the effect of pretraining-corpus scale, respiratory modality coverage, downstream data efficiency, and the contribution of whole-night sequential modeling (Appendix~\ref{app:supplementary_results}).

\section{Results}

Table~\ref{tab:main_results} summarizes pooled out-of-fold AUROC. 
Random forests based on human expert- and SOMNUS-derived sleep-summary metrics achieved comparable overall performance, with neither staging source consistently superior. The covariate-only model was particularly informative for SDB and CDH (.767 and .721), and adding covariates generally improved performance across model families. 
Within both conditions, models leveraging pretrained representations outperformed the random-forest baselines in almost all model–disorder comparisons, in line with the notion that PSG latent representations retain predictive information beyond common hypnogram-derived summaries.

\begin{table*}[!t]
  \caption{
    Pooled out-of-fold AUROC without (A) and with (B) demographic and anthropometric covariates. Best and second-best results per panel are bolded and underlined, respectively.
  }
  \label{tab:main_results}
  \centering
  \scriptsize
  \setlength{\tabcolsep}{2.0pt}
  \renewcommand{\arraystretch}{0.92}

  \begin{tabular*}{\linewidth}{
@{\extracolsep{\fill}}lccccccc@{}
}
\toprule
Method & Insomnia & SDB & CDH & CRSWD & Parasomnia & SRMD & Macro \\
\midrule

\multicolumn{8}{@{}l}{\textit{A. Without covariates}} \\
\addlinespace[1pt]

Random Forest (Expert)
  & .695 & .706 & .682 & .592 & .524 & .544 & .624 \\
Random Forest (SOMNUS)
  & .676 & .699 & .708 & .598 & .618 & .553 & .642 \\

\addlinespace[1pt]
U-Sleep
  & \underline{.726}
  & .854
  & .774
  & .625
  & \underline{.766}
  & \underline{.642}
  & \underline{.731} \\
SleepTransformer
  & \textbf{.738}
  & .874
  & \textbf{.787}
  & .616
  & \textbf{.803}
  & \textbf{.644}
  & \textbf{.744} \\

\addlinespace[1pt]
SleepFM
  & .709
  & \underline{.875}
  & .759
  & \underline{.636}
  & .726
  & .625
  & .722 \\
OSF
  & .709
  & \underline{.875}
  & .753
  & \textbf{.661}
  & .737
  & .621
  & .726 \\
Hypnos
  & .714
  & \textbf{.890}
  & \underline{.779}
  & .609
  & .748
  & .628
  & .728 \\

\midrule

\multicolumn{8}{@{}l}{\textit{B. With covariates}} \\
\addlinespace[1pt]

Random Forest (Covariates only)
  & .641 & .767 & .721 & .521 & .618 & .564 & .639 \\

\addlinespace[1pt]
Random Forest (Expert)
  & .709 & .789 & .749 & .618 & .637 & .541 & .674 \\
Random Forest (SOMNUS)
  & .682 & .777 & .759 & .558 & .701 & .550 & .671 \\

\addlinespace[1pt]
U-Sleep
  & \underline{.738}
  & .860
  & \underline{.784}
  & .617
  & \underline{.768}
  & \underline{.648}
  & \underline{.736} \\
SleepTransformer
  & \textbf{.742}
  & .874
  & \textbf{.793}
  & .604
  & \textbf{.806}
  & \textbf{.649}
  & \textbf{.745} \\

\addlinespace[1pt]
SleepFM
  & .722 & .877 & .768 & .612 & .741 & .631 & .725 \\
OSF
  & .724
  & \underline{.880}
  & .762
  & \textbf{.659}
  & .752
  & .627
  & .734 \\
Hypnos
  & .724
  & \textbf{.887}
  & .783
  & \underline{.633}
  & .730
  & .620
  & .730 \\

\bottomrule
\end{tabular*}
\end{table*}

\begin{figure*}[b!]
  \centering
  \includegraphics[width=\textwidth]
    {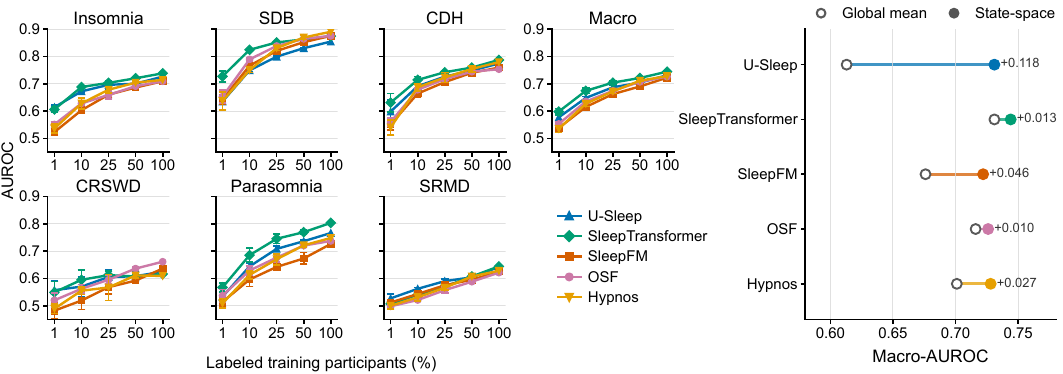}
  \caption{
     Downstream performance without covariates. Left: labeled-data efficiency; points and error bars denote mean out-of-fold AUROC and one standard deviation across participant-level draws. Right: macro-AUROC using linear probing and missingness-aware state-space modeling. 
  }
  \label{fig:data-efficiency-downstream}
\end{figure*}

The sleep-stage-supervised encoders achieved the two highest macro-AUROCs under both covariate conditions (Table~\ref{tab:main_results}). Without covariates, U-Sleep and SleepTransformer reached .731 and .744, versus .722, .726, and .728 for PSG FMs; supervised encoders led for insomnia, CDH, parasomnia, and SRMD, whereas PSG FMs led for SDB and CRSWD.
Notably, PSG FMs outperformed an EEG-only, single-cohort SleepTransformer baseline (.722--.728 versus .704). However, SleepTransformer performance increased consistently with broader supervised pretraining and modality coverage (Appendix Table~\ref{tab:st_pretraining_modality}): macro-AUROC reached .716 with multi-cohort EEG pretraining, .724 after adding EOG, EMG, and ECG, and .743 after adding respiratory effort, exceeding PSG FMs. Respiratory effort produced the largest gain for SDB (.826 $\to$ .874).
Consistently, removing respiratory groups reduced SDB AUROC by .040--.093 across all five encoders (Appendix~\ref{app:resp_ablation}), identifying respiratory coverage as a major determinant of SDB performance. 
Thus, comparisons with unimodal, single-cohort baselines may overstate the contribution of proposed architectures or pretraining objectives, as observed gains may instead largely reflect pretraining scale, cohort diversity, and modality coverage.
PSG FMs did not exhibit greater labeled-data efficiency. Across training fractions of 1\%, 10\%, 25\%, 50\%, and 100\%, the best supervised encoder exceeded the best PSG FM by .045, .039, .030, .010, and .016 macro-AUROC, respectively (Figure~\ref{fig:data-efficiency-downstream}, left; Appendix~\ref{app:data_efficiency}).
Finally, whole-night modeling improved performance for every encoder, but gains varied markedly (Figure~\ref{fig:data-efficiency-downstream}, right; Appendix~\ref{app:downstream_modeling}). SleepTransformer remained strongest under linear probing and, like OSF, declined only marginally, consistent with OSF's original evaluation \citep{Shuai2026}, whereas U-Sleep declined substantially. Thus, the accessibility of disorder information to a linear readout, and the benefit of whole-night modeling, were representation-dependent rather than consistently determined by the pretraining strategy.

\section{Conclusions}
For SWCD identification, representations from multi-cohort neural sleep scorers spanning the full physiological input set matched or outperformed those from prominent PSG FMs and exhibited greater downstream label efficiency.
The supervised encoders reused EEG/EOG-oriented preprocessing, architectures, and five-class staging objectives across modalities. Modality-specific encoders or auxiliary unimodal objectives with fewer classes may widen the gap further.
SWCD identification is closely related to sleep architecture and was evaluated in a single tertiary referral registry; additional cohorts and more distant tasks are needed to establish broader generalizability and may reveal settings in which PSG FMs offer clearer advantages.
PSG FMs nevertheless recovered much of the downstream performance of large-scale neural sleep scorers without pretraining annotations---an impressive result. Yet our findings do not support a general beyond-staging advantage for this task. Because PSG offers large-scale annotated repositories, sleep-stage-supervised representations should remain serious comparators, and advantages attributed to new objectives should be demonstrated under comparable, ideally matched, pretraining scale, diversity, and modality coverage.

\bibliography{biblio}
\bibliographystyle{icml2026}


\clearpage

\appendix

\section{Cohort and signal details}
\label{app:cohort}

\subsection{BSWR cohort characteristics}

BSWR recordings were acquired during routine clinical practice between 2002 and 2021 \cite{vanderMeer2026}. Clinical hypnograms were initially scored by trained technicians and reviewed by qualified physicians. Because the extract spans two decades of routine practice, acquisition systems and clinical scoring conventions evolved over time.

\begin{table}[hbt!]
\centering
\small
\caption{
Characteristics of the BSWR analysis cohort. Demographic and diagnostic distributions are reported at the recording level, corresponding to the unit of prediction. Diagnostic categories are non-mutually exclusive.
}
\label{tab:bswr_characteristics}
\begin{tabular}{lr}
\toprule
Characteristic & Value \\
\midrule
Patients, $n$ & 7,195 \\
Patients with repeated PSGs, $n$ (\%) & 574 (8.0) \\
PSG recordings, $n$ & 7,906 \\
Age, mean (SD), years & 48.8 (18.0) \\
Female, $n$ (\%) & 2,652 (33.5) \\
\midrule
Insomnia, $n$ (\%) & 1,131 (14.3) \\
Sleep-disordered breathing, $n$ (\%) & 5,557 (70.3) \\
Central disorders of hypersomnolence, $n$ (\%) & 800 (10.1) \\
Circadian rhythm sleep-wake disorders, $n$ (\%) & 92 (1.2) \\
Parasomnias, $n$ (\%) & 401 (5.1) \\
Sleep-related movement disorders, $n$ (\%) & 659 (8.3) \\
\midrule
No target diagnosis, $n$ (\%) & 1,056 (13.4) \\
One target diagnosis, $n$ (\%) & 5,274 (66.7) \\
Two or more target diagnoses, $n$ (\%) & 1,576 (19.9) \\
\bottomrule
\end{tabular}
\par\vspace{1mm}
\begin{minipage}{0.82\linewidth}
\footnotesize
\raggedright
\textit{Note.} ``No target diagnosis'' denotes absence of the six
modeled diagnostic groups, not a healthy-control designation.
\end{minipage}
\end{table}

\subsection{PSG signal availability}

Table~\ref{tab:bswr_signals} summarizes the signal families and canonical channel configurations used by at least one evaluated encoder. Not every listed signal was present in every PSG recording of the BSWR cohort, and individual encoders received only the channels supported by their respective input specifications as later summarized in Table \ref{tab:encoder_specs}.

\begin{table}[hbt!]
\centering
\small
\caption{
Canonical PSG signal families supplied to at least one evaluated encoder.
}
\label{tab:bswr_signals}
\begin{tabular}{p{0.14\linewidth}p{0.78\linewidth}}
\toprule
Modality & Channels and derivations \\
\midrule
EEG &
Central, frontal, and occipital channels (C3, C4, F3, F4, O1, O2, and Cz), including recordings referenced to A1/A2 or M1/M2. \\

EOG &
Left and right electro-oculography (E1 and E2), available either as pre-referenced channels or derived against the corresponding A1/A2 or M1/M2 reference. \\

EMG &
Central, upper, left, and right chin EMG, left and right tibialis or leg EMG. \\

ECG &
ECG/EKG channels \\

RESP &
Thoracic and abdominal (THX, ABD) respiratory effort; nasal pressure, cannula, or airflow; oral or thermistor airflow; snoring; oxygen saturation and pulse. \\
\bottomrule
\end{tabular}
\end{table}

\subsection{Ethical approval} The secondary use of the BSWR dataset was approved by the ethics local committee (Kantonale Ethikkommission Bern [KEK]-Nr. 2022-00415), ensuring compliance with the Human Research Act (HRA) and Ordinance on Human Research with the Exception of Clinical Trials (HRO). All methods were carried out in accordance with relevant guidelines and regulations. Data were maintained with confidentiality throughout the study.

\section{Conventional hypnogram-derived baseline specification}
\label{apd:rf}

Whole-night sleep features were calculated separately from the clinician-scored reference hypnogram and from automatic staging produced by SOMNUS, which combined predictions from the EEG- and EOG-based SLEEPYLAND sleep-staging models through soft voting \cite{DeiRossi2025}. Both hypnogram sources were converted into the same 34-feature representation: wake after sleep onset, total sleep time, sleep fragmentation index, sleep-onset latency, REM latency, the percentages of sleep assigned to N1, N2, N3, and REM, and 25 post-sleep-onset stage-dynamics features \cite{Vallat2021}.
The stage-dynamics features comprised every directed pairing among Wake, N1, N2, N3, and REM, including same-stage persistence. Each count was divided by the total number of stage transitions observed from sleep onset to the end of the recording, yielding a $5\times5$ matrix of nightly transition proportions \citep{Bechny2025}. Missing transition counts denoted an unobserved transition and were assigned zero. All other missing sleep-feature values remained missing.

The demographic and anthropometric feature set comprised age, gender, body mass index, and waist-to-height ratio. We evaluated these covariates alone and in combination with either clinician-derived or SOMNUS-derived sleep features. A separate random-forest classifier was fitted for each diagnostic category using balanced class weighting. Missing sleep-feature and covariate values were handled natively by the classifiers without imputation.

All reported predictions were obtained from the five frozen participant-grouped outer folds. Within each outer development partition, hyperparameters were selected by AUROC using five-fold stratified inner cross-validation and randomized search over the following space: 100 or 250 trees; maximum depths of 3, 5, 10, or unlimited depth; minimum split sizes of 2, 5, or 10; and minimum leaf sizes of 1, 2, or 4. Fifteen configurations were evaluated per search. The selected model was refitted on the complete outer development partition before evaluation on the held-out fold.

\section{Encoder and representation specifications}
\label{app:encoder_specs}

\begin{table}[b!]
  \caption{
    Specifications of the evaluated frozen pretrained encoders.
  }
  \label{tab:encoder_specs}
  \centering
  \scriptsize
  \setlength{\tabcolsep}{2.2pt}
  \renewcommand{\arraystretch}{1.12}
  \renewcommand{\tabularxcolumn}[1]{m{#1}}

  \begin{tabularx}{\textwidth}{
    @{}
    >{\raggedright\arraybackslash}m{0.145\textwidth}
    >{\raggedright\arraybackslash}X
    >{\centering\arraybackslash}m{0.090\textwidth}
    >{\centering\arraybackslash}m{0.055\textwidth}
    >{\raggedright\arraybackslash}m{0.250\textwidth}
    >{\centering\arraybackslash}m{0.095\textwidth}
    @{\hspace{10pt}}
    >{\centering\arraybackslash}m{0.065\textwidth}
    @{}
  }
    \toprule
    Model &
    Pretraining objective &
    Pretraining scale &
    \shortstack[c]{\#\\Params.} &
    Native BSWR inputs &
    Embedding dimension &
    Resolution \\
    \midrule

    U-Sleep &
    Sleep-stage classification &
    \shortstack[c]{$\sim$220k h\\17 cohorts} &
    15.6M &
    \shortstack[l]{
      EEG: available C3/C4/Cz,\\
      \phantom{EEG: }F3/F4, O1/O2 derivations\\
      EOG: available E1/E2\\
      \phantom{EOG: }derivations\\
      EMG: Chin/ChinL/ChinR\\
      ECG: EKG\\
      RESP: ABD/THX
    } &
    \shortstack[c]{$\mathbb{R}^{6}$\\per modality} &
    $1\,\mathrm{Hz}$ \\

    \addlinespace[4pt]

    \midrule

    SleepTransformer &
    Sleep-stage classification &
    \shortstack[c]{$\sim$220k h\\17 cohorts} &
    19.2M &
    \shortstack[l]{
      EEG: available C3/C4/Cz,\\
      \phantom{EEG: }F3/F4, O1/O2 derivations\\
      EOG: available E1/E2\\
      \phantom{EOG: }derivations\\
      EMG: Chin/ChinL/ChinR\\
      ECG: EKG\\
      RESP: ABD/THX
    } &
    \shortstack[c]{$\mathbb{R}^{128}$\\per modality} &
    $\frac{1}{30}\,\mathrm{Hz}$ \\

    \specialrule{0.8pt}{5pt}{5pt}

    SleepFM &
    Leave-one-modality-out contrastive learning &
    \shortstack[c]{432k h\\4 cohorts} &
    4.8M &
    \shortstack[l]{
      BAS: available EEG/EOG\\
      \phantom{BAS: }derivations\\
      EKG: EKG\\
      EMG: Chin/ChinL/ChinR\\
      RESP: ABD/THX, airflow,\\
      \phantom{RESP: }snoring, SpO$_2$/pulse
    } &
    \shortstack[c]{$\mathbb{R}^{128}$\\per modality} &
    $0.2\,\mathrm{Hz}$ \\

    \addlinespace[4pt]

    \midrule

    OSF &
    DINO self-distillation with channel and temporal masking &
    \shortstack[c]{$\sim$95k h\\5 cohorts} &
    85.3M &
    \shortstack[l]{
      EEG: C3-(A2/M2),\\
      \phantom{EEG: }C4-(A1/M1)\\
      EOG: E1-(A2/M2),\\
      \phantom{EOG: }E2-(A1/M1)\\
      EMG: chin and bilateral leg\\
      ECG: EKG\\
      RESP: ABD/THX, nasal pressure,\\
      \phantom{RESP: }snoring
    } &
    \shortstack[c]{$\mathbb{R}^{768}$\\fused} &
    $\frac{1}{30}\,\mathrm{Hz}$ \\

    \addlinespace[4pt]

    \midrule

    Hypnos &
    Autoregressive next-token prediction over quantized signals &
    \shortstack[c]{$\sim$135k h\\8 cohorts} &
    450.7M &
    \shortstack[l]{
      EEG: C3-(M2/A2),\\
      \phantom{EEG: }C4-(M1/A1)\\
      EOG: E1-(M2/A2),\\
      \phantom{EOG: }E2-(M1/A1)\\
      EMG: bipolar chin\\
      ECG: EKG\\
      RESP: ABD/THX
    } &
    \shortstack[c]{$\mathbb{R}^{768}$\\per stream} &
    $1\,\mathrm{Hz}$ \\

    \bottomrule
  \end{tabularx}

  \vspace{1mm}
  \begin{minipage}{0.98\linewidth}
    \scriptsize
    U-Sleep and SleepTransformer counts sum the five distinct modality-specific checkpoints used in this study; the corresponding per-checkpoint counts were 3.1M and 3.8M. The Hypnos count includes its 221.5M-parameter RQ-Transformer and five complete waveform tokenizers.
    Reported cohort counts follow the granularity used in the source publications and do not represent harmonized counts of acquisition sites. Some named cohorts aggregate recordings from multiple centers (e.g., SleepFM's BioSerenity source spans 240 US facilities), whereas SLEEPYLAND counts the CNC, DHC, and SSC subcohorts of MNC separately. These values should therefore be interpreted as reported pretraining data sources or partitions rather than directly comparable measures of site diversity. 
  \end{minipage}
\end{table}

Table~\ref{tab:encoder_specs} summarizes the pretraining objectives, reported pretraining scales, frozen parameter counts, per modality channels employed, and dimensions and resolutions of the representations leveraged by downstream modeling. 
Parameter counts cover the complete frozen extraction bundle, characterizing the model checkpoint footprint. Pretraining scales follow the original publications; when the number of pretraining hours was not reported, it was estimated from the reported number of recordings by assuming a mean duration of eight hours per recording. BSWR channels were mapped to each encoder's native input specification; where specific EEG or EOG references were required, existing derivations were retained or reconstructed from their component electrodes. 

\subsection{Common extraction and input handling}

All pretrained checkpoints remained frozen. We preserved the preprocessing prescribed for each checkpoint. Signal resampling, normalization, channel mapping, and signal-level missing-input handling therefore followed the corresponding model specification, as detailed below.

Extracted representation sequences were retained at their native temporal resolutions. No additional temporal averaging, interpolation, or resampling was applied before downstream modeling. Missing representation streams were identified through modality masks, while sequences of different duration were right-padded within each batch and accompanied by temporal-validity masks, so that temporal pooling could eventually exclude them.

\subsection{Supervised pretrained models}
\label{app:supervised_encoders}

\paragraph{Pretraining and preprocessing}

The SLEEPYLAND models were pretrained on approximately 220,000 hours of expert-annotated PSG from 17 NSRR cohorts \citep{DeiRossi2025}. Each modality-specific checkpoint was trained on recordings containing the corresponding physiological signal and sleep-stage annotations. Signals were resampled to 128~Hz and normalized independently by recording and channel by robust scaling. Values were clipped at 20 interquartile ranges from the median. Signal intervals outside the scored hypnogram were excluded, as were epochs labeled as movement or unknown.
The same multi-cohort sampling framework was used across architectures and modalities. Datasets were sampled using an equal mixture of uniform-over-dataset and dataset-size-proportional sampling. After selecting a recording, one eligible channel derivation and a 35-epoch segment were sampled. Segment location was conditioned on a randomly selected sleep stage to increase exposure to less prevalent classes. Data augmentation replaced randomly selected signal intervals with Gaussian noise. The EMG, ECG, and RESP models considered in this study followed the same preprocessing, sampling, augmentation, and architecture-specific optimization framework as the released EEG and EOG checkpoints.

\paragraph{Representation extraction}

All encoder parameters were frozen. Each eligible BSWR channel (reported in Table~\ref{tab:encoder_specs}) was processed independently using the corresponding unimodal checkpoint, producing one temporal representation for each recording, modality, and available channel.

\paragraph{U-Sleep}

U-Sleep is a fully convolutional encoder--decoder with 12 resolution levels \citep{Perslev2021}. At each encoder level, temporal convolution, ELU activation, and batch normalization are followed by max pooling, progressively increasing the receptive field while reducing temporal resolution. The decoder restores the original resolution through upsampling and convolution, concatenating features from the corresponding encoder levels through skip connections. The resulting high-resolution feature map is projected by a pointwise convolution onto six learned feature channels.
The U-Sleep segment classifier first averages this six-channel representation over a configurable temporal interval and then applies two five-channel convolutional layers to produce sleep-stage probabilities. For representation extraction, the averaging interval was set to one second (128 samples), and its six-dimensional output was retained before the five-class classification layers. Each input channel therefore produced a $T \times 6$ representation, where $T$ is the scored recording duration in seconds. The fully convolutional architecture allowed complete recordings to be encoded in a single forward pass.

\paragraph{SleepTransformer}

SleepTransformer processes sequences of 35 consecutive 30-second epochs using two hierarchical transformer stages \citep{Phan2022sleeptransformer}. Each epoch is represented by 29 spectral frames and 129 frequency bins. After removing the zero-frequency bin, every spectral frame forms a 128-dimensional token for a single-channel input. Four eight-head transformer encoder blocks model relationships among these tokens, after which learned attention pooling produces one 128-dimensional representation per epoch.
A second stack of four eight-head transformer encoder blocks contextualizes the epoch representations across the 35-epoch sequence, corresponding to 17.5 minutes of inter-epoch context. The contextualized state for each epoch would ordinarily pass through two 1,024-unit fully connected layers and the five-class sleep-stage classifier. We instead retained the 128-dimensional output of the sequence transformer before this classification module. The contextualized epoch representations yielded a $T \times 128$ sequence at 30-second resolution. Each representation therefore incorporates both within-epoch spectral information and intermediate-range temporal context.

\subsection{Self-supervised pretrained models}
\label{app:self_supervised_encoders}

\paragraph{SleepFM}

SleepFM uses one shared encoder across four broad modality groups: BAS, comprising EEG and EOG; EKG; EMG; and RESP \citep{Thapa2026}. Signals were resampled to 128~Hz and standardized over each recording. Every available channel was divided into five-second windows and mapped by a six-layer convolutional tokenizer to 128-dimensional tokens. Attention-based channel pooling then combined the available channels within each modality, after which a temporal transformer contextualized the modality-specific sequence over complete five-minute intervals. During pretraining, leave-one-modality-out contrastive learning aligned the temporally pooled representation of each modality with the average representation of the remaining modalities from the same interval.

For BSWR, canonical channel names were matched to the released modality-specific allowlists, and all accepted channels within each group were processed jointly. We retained the contextualized five-second sequence preceding temporal pooling, rather than SleepFM's pooled interval representation or task-specific prediction head.

\paragraph{OSF}

OSF adapts DINO teacher-student self-distillation to a fixed 12-channel PSG montage \citep{Oquab2023,Shuai2026}. Pretraining views were generated by randomly masking half of the channels and subsequently masking temporal blocks covering between 30\% and 60\% of each example. The released vision transformer processes 30-second epochs sampled at 64~Hz. A convolutional patchifier groups four adjacent channel positions over one-second intervals, yielding 90 channel-time tokens per epoch. A prepended CLS token integrates information across these tokens through 12 transformer layers. We use the OSF-Base version, i.e., the best-performing released checkpoint.

BSWR signals were mapped to the prescribed montage using direct channels or explicitly supported derivations; unresolved positions remained zero, consistent with OSF's missing-channel training. Observed signals were standardized over the recording, resampled to 64~Hz, and clipped to $[-6,6]$. We retained the final 768-dimensional CLS state before the self-distillation projection head, producing one fused representation per complete 30-second epoch. 

\paragraph{Hypnos}

Hypnos first converts eight physiological streams into discrete residual-quantization codes using five causal waveform tokenizers shared across related signal types \citep{Carter2026}. C3 and C4 share an EEG tokenizer, E1 and E2 share an EOG tokenizer, and abdominal and thoracic effort share a respiratory tokenizer; ECG and chin EMG use separate tokenizers. Each tokenizer emits one residual-code stack per second. A causal RQ-Transformer then alternates temporal attention within streams and cross-modal attention among simultaneous streams before predicting subsequent codes across residual depth. Random modality grouping during pretraining exposes the model to varying input configurations. We use the Hypnos-Base version, i.e., the best-performing released checkpoint.

We retained the released channel resolution, re-referencing, filtering, rolling normalization, and amplitude-compression procedures. EEG, EOG, ECG, and chin EMG were supplied at 128~Hz, whereas abdominal and thoracic effort were supplied at 32~Hz. Missing streams were omitted through the native modality mask. For every available stream, we extracted the 768-dimensional temporal-transformer state before residual-depth prediction. These one-second representations retain their stream identity while already incorporating causal temporal and cross-modal context.

\section{Downstream modeling and evaluation}
\label{app:downstream}

\subsection{Representation handling and alternative-channel sampling}

The downstream loader preserved the stream organization, feature dimension, and temporal resolution defined by each encoder. Streams from the same recording were temporally aligned and right-padded to the longest sequence within each minibatch. A temporal-validity mask identified the contiguous prefix of non-padded positions, while a separate stream-availability mask indicated whether each complete physiological stream was present. Missing streams were represented by zero-valued placeholders. No additional temporal interpolation, resampling, or aggregation was applied after representation extraction.

U-Sleep and SleepTransformer could provide multiple independently encoded channels for the same physiological modality. During training, one available channel was selected for each recording and modality using a deterministic function of the experiment seed, outer fold, recording identifier, modality, and epoch. Channel configurations therefore varied across epochs while remaining reproducible. Outer-fold inference was repeated over ten deterministic channel configurations, and the resulting disorder probabilities were averaged for each recording and label. Representations with an encoder-prescribed stream organization were supplied directly without downstream channel sampling.

The temporal-validity mask was used for sequence reversal and whole-night pooling. Because valid tokens always formed a contiguous prefix and padding was restricted to the right, padded positions could not affect earlier valid outputs of the causal recurrence. 

\subsection{Mask-aware Coupled Mamba}

Let \(\mathbf{r}_t^m\in\mathbb{R}^{F_m}\) denote the frozen representation of stream \(m\) at time \(t\). Because the encoders produced different feature dimensions \(F_m\), each stream was projected independently to a common latent width:

\[
\mathbf{z}_{0,t}^m
=
\mathbf{W}_{\mathrm{proj}}^m\mathbf{r}_t^m
+
\mathbf{b}_{\mathrm{proj}}^m,
\qquad
\mathbf{z}_{0,t}^m\in\mathbb{R}^{d_{\mathrm{model}}}.
\]

Coupled Mamba blocks were arranged as pre-normalized residual layers. At layer \(\ell\), a normalized copy of each stream was supplied to the block, while the unchanged layer input was retained as the residual path:

\[
\widetilde{\mathbf{z}}_{\ell,t}^m
=
\operatorname{Norm}_{\ell,m}
\left(
\mathbf{z}_{\ell,t}^m
\right).
\]

Every stream retained separate input, depthwise-convolution, gating, state-space, readout, and output projections. A stream-specific input projection produced parallel content and gate branches,

\[
\left[
\mathbf{x}_t^m,
\mathbf{g}_t^m
\right]
=
\mathbf{W}_{\mathrm{in}}^m
\widetilde{\mathbf{z}}_{\ell,t}^m,
\]

after which the content branch underwent a depthwise causal convolution. Selective state-space projections then produced an input-dependent transition \(\mathbf{P}_t^m\), update \(\mathbf{U}_t^m\), and readout parameters for each stream.

Following the efficient Coupled Mamba recurrence, in which each
modality-specific state is conditioned on the aggregate previous states
\citep{Li2024}, we implemented a normalized, mask-aware shared-state variant. For stream-availability indicator \(a_m\in\{0,1\}\) and number of available streams \(M_{\mathrm{avail}}=\sum_m a_m,\) the shared transition and update were

\[
\overline{\mathbf{P}}_t
=
\frac{1}{M_{\mathrm{avail}}}
\sum_m a_m\mathbf{P}_t^m,
\qquad
\overline{\mathbf{U}}_t
=
\frac{1}{M_{\mathrm{avail}}}
\sum_m a_m\mathbf{U}_t^m.
\]

Starting from a zero state, the shared recurrent trajectory evolved as

\[
\overline{\mathbf{H}}_t
=
\overline{\mathbf{P}}_t
\odot
\overline{\mathbf{H}}_{t-1}
+
\overline{\mathbf{U}}_t,
\]

where \(\odot\) denotes element-wise multiplication. Each stream subsequently
combined its own current transition and update with the preceding shared state,

\[
\mathbf{H}_t^m
=
\mathbf{P}_t^m
\odot
\overline{\mathbf{H}}_{t-1}
+
\mathbf{U}_t^m.
\]

Only \(\overline{\mathbf{H}}_t\) was propagated recurrently. The transient stream-specific state \(\mathbf{H}_t^m\) was converted into an output using its input-dependent readout. As in standard Mamba, this recurrent readout was combined with a learned direct contribution from the current locally convolved content, modulated by the parallel gate, and projected back to \(d_{\mathrm{model}}\). The projected block output was then added to the unchanged layer input through the outer residual connection.

\[
\mathbf{z}_{\ell+1,t}^m
=
\mathbf{z}_{\ell,t}^m
+
a_m\mathbf{o}_{\ell,t}^m.
\]

The resulting sequences therefore retained their original stream identity and temporal resolution while incorporating information from the shared multimodal history. Unavailable streams were excluded from shared-state averaging and their block outputs were zeroed.

In bidirectional profiles, independent parameter sets processed the valid sequence in forward and reverse temporal order. Reverse outputs were restored to the original order and averaged with the forward outputs before the residual update. For OSF, whose upstream encoder already produced one fused representation sequence, the same formulation reduced to a single-stream selective state-space recurrence; raw-signal availability had already been resolved during OSF preprocessing.

\subsection{Temporal pooling, covariates, and classification}

Following whole-night processing, each stream was summarized independently using masked mean pooling, its last valid state, or additive-attention pooling. For attention pooling, each stream had a separate scorer, mirroring the implementation of \cite{Phan2022sleeptransformer, Bahdanau2014}:

\[
e_t^m
=
\mathbf{v}_m^\top
\tanh\!\left(
\mathbf{W}_m\mathbf{y}_t^m+\mathbf{b}_m
\right),
\qquad
\alpha_t^m
=
\frac{\exp(e_t^m)}
{\sum_{s\in\mathcal{V}}\exp(e_s^m)},
\qquad
\mathbf{p}^m
=
\sum_{t\in\mathcal{V}}
\alpha_t^m\mathbf{y}_t^m ,
\]

where \(\mathcal{V}\) contains the valid, non-padded time points. The pooled stream representations were concatenated and normalized, with unavailable streams contributing zero-valued vectors.

We evaluated representation-only and with-covariate conditions. The covariate set comprised age, gender, body mass index, waist-to-height ratio, and indicators for missing anthropometric values. Age, body mass index, and waist-to-height ratio were standardized using statistics fitted exclusively to the corresponding outer-fold training partition. Missing anthropometric values were assigned zero after standardization and accompanied by separate missingness indicators. Covariates were appended only after temporal pooling and therefore did not influence the recurrent sequence representation.

The resulting features were passed to a two-layer multilayer perceptron. Its hidden layer was followed by layer normalization, GELU activation, and dropout. The final layer produced one logit for each of the six diagnostic categories.

\subsection{Cross-validation, architecture selection, and optimization}

We used five fixed participant-grouped outer folds shared across all methods, assigning every recording from a patient to the same fold. Fold construction approximately balanced the total number of recordings and diagnosis-specific positive counts. Within each outer development partition, 15\% of development patients were assigned to a fixed participant-grouped validation subset; these assignments were reused across encoders, architecture configurations, and covariate conditions.

Architecture selection was performed independently within each outer development partition and separately for every encoder under the representation-only and covariate-augmented conditions. We evaluated prespecified compact configurations spanning model widths of 16 and 32, state dimensions of 8 and 16, one or two Coupled Mamba layers, unidirectional or bidirectional processing, and mean, last-state, or additive-attention pooling. Within each outer run, configurations were ranked by macro-AUROC on that run's fixed validation subset. The corresponding early-stopped checkpoint of the selected configuration was then evaluated once on the held-out outer fold. Consequently, the selected architecture could differ across outer folds, and the pooled out-of-fold predictions estimate the performance of the complete training-and-selection procedure.

Across configurations, the causal-convolution width was fixed to \(d_{\mathrm{conv}}=4\), the expansion factor to 2, and classifier dropout to 0.1; layer normalization was used throughout. Models were optimized jointly across the six labels using unweighted binary cross-entropy and Adam with a learning rate of \(5\times10^{-4}\), no weight decay, and minibatches of four recordings. Training continued for at most 100 epochs, with early stopping after ten consecutive epochs without improvement in validation loss; the checkpoint attaining the lowest validation loss was retained. Models were implemented in Python 3.10 using PyTorch 2.2.0 with CUDA 12.1, \texttt{mamba-ssm} 1.2.2, and \texttt{causal-conv1d} 1.4.0, and trained on an NVIDIA H200 GPU.

\section{Supplementary results}
\label{app:supplementary_results}

\paragraph{SleepTransformer pretraining and modality conditions}\label{app:modality_coverage}

Using the same fixed downstream architecture, we compared four SleepTransformer \cite{Phan2022sleeptransformer} conditions. The single-cohort EEG-only condition used an encoder pretrained exclusively on the large SHHS cohort, whereas the multi-cohort EEG-only condition used the corresponding encoder pretrained across the 17 NSRR cohorts considered in SLEEPYLAND \cite{DeiRossi2025}. The multi-cohort core condition additionally incorporated separately pretrained EOG, EMG, and ECG encoders, while the respiratory-expanded condition further incorporated ABD and THX respiratory-effort encoders. The two EEG-only conditions examined the influence of pretraining-corpus scale and diversity while holding the encoder architecture and downstream modality fixed. The remaining comparisons quantified the contributions of broader modality coverage and, particularly, respiratory effort.

\begin{table}[hbt!]
\caption{
Performance of SleepTransformer downstream modeling across pretraining and modality-coverage conditions. Values are pooled out-of-fold AUROC without covariates. Best and second-best results are bolded and underlined.
}
\label{tab:st_pretraining_modality}
\centering
\scriptsize
\setlength{\tabcolsep}{2.2pt}
\renewcommand{\arraystretch}{0.95}

\begin{tabular*}{\linewidth}{
@{\extracolsep{\fill}}lccccccc@{}
}
\toprule
Pretraining and inputs
& Insomnia & SDB & CDH & CRSWD & Parasomnia & SRMD & Macro \\
\midrule

Single-cohort; EEG
  & .700
  & .786
  & .753
  & \underline{.617}
  & .764
  & .603
  & .704 \\

Multi-cohort; EEG
  & .709
  & .797
  & .759
  & \textbf{.631}
  & \underline{.795}
  & .603
  & .716 \\

Multi-cohort; EEG/EOG/EMG/ECG
  & \underline{.712}
  & \underline{.826}
  & \underline{.778}
  & .604
  & .792
  & \underline{.630}
  & \underline{.724} \\

Multi-cohort; EEG/EOG/EMG/ECG/RESP
  & \textbf{.737}
  & \textbf{.874}
  & \textbf{.787}
  & .609
  & \textbf{.800}
  & \textbf{.649}
  & \textbf{.743} \\

\bottomrule
\end{tabular*}

\vspace{1mm}
\begin{minipage}{0.98\linewidth}
\tiny
All ablation conditions used the same fixed downstream architecture configuration. The respiratory-expanded results therefore differ slightly from the primary results, for which the downstream architecture was selected independently within each outer development partition.
\end{minipage}
\end{table}

\paragraph{Respiratory-input ablation}\label{app:resp_ablation}

To harmonize physiological coverage as far as permitted by each encoder, we repeated the analyses using a common non-respiratory core comprising EEG, EOG, chin EMG, and ECG. Respiratory coverage represented the principal input mismatch, ranging from respiratory effort alone to combinations of effort, airflow, oxygen saturation, pulse, and snoring. U-Sleep and SleepTransformer omitted their RESP group; SleepFM omitted its RESP group; OSF retained only EEG, EOG, chin EMG, and ECG, thereby also excluding leg EMG; and Hypnos omitted ABD and THX. Because each encoder retained its own preprocessing, channel handling, and representation organization, this constituted signal-family-level harmonization rather than exact input matching.

\begin{table}[hbt!]
\caption{
Pooled out-of-fold AUROC under common-core input coverage, without (A) and with (B) demographic and anthropometric covariates. Best and second-best available results per panel are bolded and underlined. 
}
\label{tab:resp_common_core}
\centering
\scriptsize
\setlength{\tabcolsep}{2.1pt}
\renewcommand{\arraystretch}{0.94}

\begin{tabular*}{\linewidth}{
@{\extracolsep{\fill}}lccccccc@{}
}
\toprule
Method & Insomnia & SDB & CDH & CRSWD & Parasomnia & SRMD & Macro \\
\midrule

\multicolumn{8}{@{}l}{\textit{A. Without covariates}} \\
\addlinespace[1pt]

U-Sleep
  & \underline{.712}
  & .814
  & \underline{.772}
  & \underline{.629}
  & \underline{.751}
  & \underline{.635}
  & \underline{.719} \\
SleepTransformer
  & \textbf{.718}
  & \textbf{.827}
  & \textbf{.783}
  & .608
  & \textbf{.802}
  & \textbf{.639}
  & \textbf{.730} \\
SleepFM
  & .686 & .782 & .751 & .608 & .689 & .594 & .685 \\
OSF
  & .697 & .810 & .748 & \textbf{.637} & .709 & .603 & .701 \\
Hypnos
  & .702 & \underline{.816} & .758 & .620 & .715 & .597 & .701 \\

\midrule

\multicolumn{8}{@{}l}{\textit{B. With covariates}} \\
\addlinespace[1pt]

U-Sleep
  & \underline{.728}
  & \underline{.827}
  & \textbf{.788}
  & \textbf{.642}
  & \underline{.774}
  & \underline{.637}
  & \textbf{.733} \\
SleepTransformer
  & \textbf{.734}
  & \textbf{.837}
  & \underline{.786}
  & .574
  & \textbf{.806}
  & \textbf{.647}
  & \underline{.731} \\
SleepFM
  & .700 & .803 & .768 & .603 & .715 & .614 & .701 \\
OSF
  & .712 & .820 & .771 & \underline{.633} & .706 & .599 & .707 \\
Hypnos
  & .707 & .809 & .766 & .587 & .729 & .611 & .701 \\

\bottomrule
\end{tabular*}
\end{table}

Under common-core coverage (Table \ref{tab:resp_common_core}), the supervised encoders remained strongest overall: SleepTransformer achieved the highest macro-AUROC without covariates (.730), while U-Sleep ranked first with covariates (.733). Unlike under full input coverage, SleepTransformer also achieved the highest SDB AUROC in both panels (.827 and .837). Removing respiratory groups reduced SDB AUROC by .040--.093 across the five encoders, whereas changes for the other disorders were smaller and less consistent. Overall, the common-core comparison indicates that much of the SDB advantage observed with the full PSG FM inputs arose from respiratory coverage.

\begin{table}[hbt!]
\caption{
SleepFM SDB performance across full, non-respiratory-core, RESP-only,
and core-plus-respiratory-subset conditions.
}
\label{tab:sleepfm_resp_decomposition}
\centering
\scriptsize
\setlength{\tabcolsep}{5pt}
\renewcommand{\arraystretch}{0.94}

\begin{tabular}{lc}
\toprule
SleepFM inputs & SDB AUROC \\
\midrule
All modality groups                    & \textbf{.879} \\
BAS/ECG/EMG only                       & .769 \\
RESP group only                        & .869 \\
BAS/ECG/EMG + respiratory effort       & .865 \\
BAS/ECG/EMG + airflow                  & .864 \\
BAS/ECG/EMG + oximetry/pulse           & .847 \\
BAS/ECG/EMG + snoring                  & .776 \\
\bottomrule
\end{tabular}
\end{table}

To examine this effect further, we evaluated the complete SleepFM RESP group and its constituent signal families (Table~\ref{tab:sleepfm_resp_decomposition}). The RESP-group-only condition excluded BAS, ECG, and EMG; conversely, each respiratory-subset condition retained these non-respiratory groups and restricted only the RESP inputs to the indicated family. All conditions used the same fixed downstream architecture and excluded covariates.
Removing the RESP group reduced SDB AUROC from .879 to .769, whereas the RESP group alone retained nearly all full-model performance (.869). Adding respiratory effort or airflow to the non-respiratory core recovered AUROC to .865 and .864, respectively, followed by oximetry/pulse (.847); snoring provided little improvement (.776). Thus, the SDB gain from respiratory coverage was driven primarily by effort and airflow, while the complete RESP group achieved the highest performance.

\paragraph{Downstream data efficiency}\label{app:data_efficiency}

All data-efficiency analyses used the representation-only condition, without appending demographic or anthropometric covariates to the downstream model. Within each outer fold, downstream training was restricted to 1\%, 10\%, 25\%, 50\%, or 100\% of the training participants. Participant subsets were selected to approximate the target recording count and diagnosis distribution while retaining both classes for every disorder. Identical subsets were used across encoders.
To limit dependence on a particular sample, we used five mutually disjoint draws within the 1\% and 10\% conditions, four at 25\%, and two at 50\%. The complete outer-fold training partition defined the single 100\% condition. 

Under each encoder's full physiological input coverage, SleepTransformer achieved the highest macro-AUROC at every training fraction (Table~\ref{tab:data_efficiency_full}). Its advantage over the strongest PSG FM was .045, .039, .030, .010, and .016 at 1\%, 10\%, 25\%, 50\%, and 100\%, respectively. U-Sleep showed a similar pattern, exceeding the strongest PSG FM at four of five fractions and trailing it by only .007 at 50\%. The supervised advantage was most consistent for insomnia, CDH, and parasomnia. For SDB, differences narrowed with increasing training data, with PSG FMs slightly leading at 50\% and 100\%. SRMD remained close to chance at the smallest fractions and separated more clearly as training data increased.

The common non-respiratory configuration produced the same overall pattern (Table~\ref{tab:data_efficiency_core}). SleepTransformer exceeded the strongest corresponding PSG FM by .052, .041, .041, .040, and .029 macro-AUROC across the five fractions, while U-Sleep remained above every PSG FM throughout. Supervised representations generally performed best across disorders, particularly for SDB, CDH, and parasomnia. CRSWD showed the greatest variation across participant draws and was the principal exception at full data, where OSF achieved the highest AUROC.

\begin{table*}[hbt!]
\caption{
Downstream data efficiency under each encoder's full physiological input coverage, without demographic and anthropometric covariates. Reduced-data values are mean \(\pm\) standard deviation across participant draws. Best and second-best mean AUROCs within each training fraction and outcome are bolded and underlined, respectively.
}
\label{tab:data_efficiency_full}
\centering
\scriptsize
\setlength{\tabcolsep}{1.1pt}
\renewcommand{\arraystretch}{0.94}

\begin{tabular*}{\linewidth}{
@{\extracolsep{\fill}}llccccccc@{}
}
\toprule
Method & Train (\%) & Insomnia & SDB & CDH & CRSWD & Parasomnia & SRMD & Macro \\
\midrule

U-Sleep
  & 1
  & $\mathbf{.614}{\scriptstyle\pm.008}$
  & $.635{\scriptstyle\pm.011}$
  & $\underline{.598}{\scriptstyle\pm.042}$
  & $\mathbf{.557}{\scriptstyle\pm.033}$
  & $.533{\scriptstyle\pm.016}$
  & $\mathbf{.527}{\scriptstyle\pm.016}$
  & $\underline{.577}{\scriptstyle\pm.016}$ \\
  & 10
  & $\underline{.672}{\scriptstyle\pm.003}$
  & $.749{\scriptstyle\pm.006}$
  & $\underline{.691}{\scriptstyle\pm.006}$
  & $\underline{.570}{\scriptstyle\pm.036}$
  & $\underline{.644}{\scriptstyle\pm.011}$
  & $\mathbf{.562}{\scriptstyle\pm.004}$
  & $\underline{.648}{\scriptstyle\pm.006}$ \\
  & 25
  & $\underline{.696}{\scriptstyle\pm.016}$
  & $.799{\scriptstyle\pm.003}$
  & $\underline{.726}{\scriptstyle\pm.008}$
  & $\underline{.605}{\scriptstyle\pm.027}$
  & $\underline{.708}{\scriptstyle\pm.010}$
  & $\mathbf{.590}{\scriptstyle\pm.007}$
  & $\underline{.687}{\scriptstyle\pm.006}$ \\
  & 50
  & $.700{\scriptstyle\pm.004}$
  & $.830{\scriptstyle\pm.003}$
  & $.748{\scriptstyle\pm.011}$
  & $.608{\scriptstyle\pm.017}$
  & $\underline{.736}{\scriptstyle\pm.005}$
  & $\underline{.604}{\scriptstyle\pm.005}$
  & $.704{\scriptstyle\pm.000}$ \\
  & 100
  & \underline{.726}
  & .854
  & .774
  & .625
  & \underline{.766}
  & \underline{.642}
  & \underline{.731} \\

\addlinespace[1pt]
SleepTransformer
  & 1
  & $\underline{.606}{\scriptstyle\pm.010}$
  & $\mathbf{.727}{\scriptstyle\pm.020}$
  & $\mathbf{.631}{\scriptstyle\pm.034}$
  & $\underline{.549}{\scriptstyle\pm.039}$
  & $\mathbf{.567}{\scriptstyle\pm.016}$
  & $.507{\scriptstyle\pm.015}$
  & $\mathbf{.598}{\scriptstyle\pm.012}$ \\
  & 10
  & $\mathbf{.688}{\scriptstyle\pm.005}$
  & $\mathbf{.824}{\scriptstyle\pm.005}$
  & $\mathbf{.714}{\scriptstyle\pm.013}$
  & $\mathbf{.595}{\scriptstyle\pm.036}$
  & $\mathbf{.685}{\scriptstyle\pm.026}$
  & $.540{\scriptstyle\pm.009}$
  & $\mathbf{.674}{\scriptstyle\pm.010}$ \\
  & 25
  & $\mathbf{.703}{\scriptstyle\pm.004}$
  & $\mathbf{.851}{\scriptstyle\pm.002}$
  & $\mathbf{.742}{\scriptstyle\pm.003}$
  & $\mathbf{.612}{\scriptstyle\pm.015}$
  & $\mathbf{.745}{\scriptstyle\pm.018}$
  & $.570{\scriptstyle\pm.013}$
  & $\mathbf{.704}{\scriptstyle\pm.005}$ \\
  & 50
  & $\mathbf{.720}{\scriptstyle\pm.002}$
  & $.862{\scriptstyle\pm.001}$
  & $\mathbf{.759}{\scriptstyle\pm.001}$
  & $.607{\scriptstyle\pm.015}$
  & $\mathbf{.769}{\scriptstyle\pm.009}$
  & $\mathbf{.607}{\scriptstyle\pm.009}$
  & $\mathbf{.721}{\scriptstyle\pm.002}$ \\
  & 100
  & \textbf{.738}
  & .874
  & \textbf{.787}
  & .616
  & \textbf{.803}
  & \textbf{.644}
  & \textbf{.744} \\

\addlinespace[1pt]
SleepFM
  & 1
  & $.523{\scriptstyle\pm.012}$
  & $.643{\scriptstyle\pm.017}$
  & $.551{\scriptstyle\pm.020}$
  & $.482{\scriptstyle\pm.028}$
  & $.514{\scriptstyle\pm.015}$
  & $\underline{.512}{\scriptstyle\pm.007}$
  & $.538{\scriptstyle\pm.009}$ \\
  & 10
  & $.603{\scriptstyle\pm.010}$
  & $.766{\scriptstyle\pm.005}$
  & $.664{\scriptstyle\pm.013}$
  & $.519{\scriptstyle\pm.032}$
  & $.596{\scriptstyle\pm.025}$
  & $\underline{.543}{\scriptstyle\pm.007}$
  & $.615{\scriptstyle\pm.007}$ \\
  & 25
  & $.660{\scriptstyle\pm.009}$
  & $.820{\scriptstyle\pm.011}$
  & $.706{\scriptstyle\pm.009}$
  & $.568{\scriptstyle\pm.024}$
  & $.642{\scriptstyle\pm.009}$
  & $\underline{.575}{\scriptstyle\pm.006}$
  & $.662{\scriptstyle\pm.006}$ \\
  & 50
  & $.686{\scriptstyle\pm.000}$
  & $.852{\scriptstyle\pm.005}$
  & $.740{\scriptstyle\pm.003}$
  & $.592{\scriptstyle\pm.003}$
  & $.673{\scriptstyle\pm.021}$
  & $.595{\scriptstyle\pm.002}$
  & $.690{\scriptstyle\pm.005}$ \\
  & 100
  & .709
  & \underline{.875}
  & .759
  & \underline{.636}
  & .726
  & .625
  & .722 \\

\addlinespace[1pt]
OSF
  & 1
  & $.551{\scriptstyle\pm.010}$
  & $\underline{.653}{\scriptstyle\pm.023}$
  & $.560{\scriptstyle\pm.014}$
  & $.520{\scriptstyle\pm.027}$
  & $\underline{.537}{\scriptstyle\pm.023}$
  & $.498{\scriptstyle\pm.004}$
  & $.553{\scriptstyle\pm.007}$ \\
  & 10
  & $.628{\scriptstyle\pm.006}$
  & $\underline{.789}{\scriptstyle\pm.005}$
  & $.678{\scriptstyle\pm.007}$
  & $.562{\scriptstyle\pm.016}$
  & $.629{\scriptstyle\pm.017}$
  & $.521{\scriptstyle\pm.006}$
  & $.635{\scriptstyle\pm.005}$ \\
  & 25
  & $.659{\scriptstyle\pm.014}$
  & $\underline{.838}{\scriptstyle\pm.004}$
  & $.717{\scriptstyle\pm.006}$
  & $.595{\scriptstyle\pm.016}$
  & $.676{\scriptstyle\pm.015}$
  & $.557{\scriptstyle\pm.011}$
  & $.674{\scriptstyle\pm.003}$ \\
  & 50
  & $.691{\scriptstyle\pm.001}$
  & $\underline{.864}{\scriptstyle\pm.003}$
  & $.744{\scriptstyle\pm.003}$
  & $\mathbf{.636}{\scriptstyle\pm.005}$
  & $.720{\scriptstyle\pm.003}$
  & $.588{\scriptstyle\pm.003}$
  & $.707{\scriptstyle\pm.002}$ \\
  & 100
  & .709
  & \underline{.875}
  & .753
  & \textbf{.661}
  & .737
  & .621
  & .726 \\

\addlinespace[1pt]
Hypnos
  & 1
  & $.536{\scriptstyle\pm.016}$
  & $.637{\scriptstyle\pm.033}$
  & $.540{\scriptstyle\pm.028}$
  & $.489{\scriptstyle\pm.017}$
  & $.509{\scriptstyle\pm.017}$
  & $.500{\scriptstyle\pm.012}$
  & $.535{\scriptstyle\pm.013}$ \\
  & 10
  & $.628{\scriptstyle\pm.020}$
  & $.750{\scriptstyle\pm.009}$
  & $.687{\scriptstyle\pm.014}$
  & $.555{\scriptstyle\pm.014}$
  & $.613{\scriptstyle\pm.015}$
  & $.531{\scriptstyle\pm.013}$
  & $.627{\scriptstyle\pm.010}$ \\
  & 25
  & $.678{\scriptstyle\pm.012}$
  & $.830{\scriptstyle\pm.009}$
  & $.723{\scriptstyle\pm.018}$
  & $.567{\scriptstyle\pm.047}$
  & $.670{\scriptstyle\pm.020}$
  & $.566{\scriptstyle\pm.008}$
  & $.672{\scriptstyle\pm.007}$ \\
  & 50
  & $\underline{.704}{\scriptstyle\pm.002}$
  & $\mathbf{.869}{\scriptstyle\pm.007}$
  & $\underline{.754}{\scriptstyle\pm.011}$
  & $\underline{.611}{\scriptstyle\pm.009}$
  & $.721{\scriptstyle\pm.007}$
  & $\mathbf{.607}{\scriptstyle\pm.001}$
  & $\underline{.711}{\scriptstyle\pm.001}$ \\
  & 100
  & .714
  & \textbf{.890}
  & \underline{.779}
  & .609
  & .748
  & .628
  & .728 \\

\bottomrule
\end{tabular*}
\end{table*}


\begin{table*}[hbt!]
\caption{
Downstream data efficiency under non-respiratory core, without demographic and anthropometric covariates. Reduced-data values are mean \(\pm\) standard deviation across participant-level draws. Best and second-best reported mean AUROCs within each training fraction and outcome are bolded and underlined, respectively.
}
\label{tab:data_efficiency_core}
\centering
\scriptsize
\setlength{\tabcolsep}{1.1pt}
\renewcommand{\arraystretch}{0.94}

\begin{tabular*}{\linewidth}{
@{\extracolsep{\fill}}llccccccc@{}
}
\toprule
Method & Train (\%) & Insomnia & SDB & CDH & CRSWD & Parasomnia & SRMD & Macro \\
\midrule

U-Sleep
  & 1
  & $\mathbf{.605}{\scriptstyle\pm.014}$
  & $.614{\scriptstyle\pm.005}$
  & $\underline{.598}{\scriptstyle\pm.046}$
  & $\underline{.542}{\scriptstyle\pm.022}$
  & $\underline{.546}{\scriptstyle\pm.006}$
  & $\underline{.507}{\scriptstyle\pm.014}$
  & $\underline{.569}{\scriptstyle\pm.012}$ \\
  & 10
  & $\underline{.662}{\scriptstyle\pm.004}$
  & $.715{\scriptstyle\pm.006}$
  & $\underline{.697}{\scriptstyle\pm.007}$
  & $\underline{.572}{\scriptstyle\pm.044}$
  & $\underline{.641}{\scriptstyle\pm.016}$
  & $\underline{.546}{\scriptstyle\pm.011}$
  & $\underline{.639}{\scriptstyle\pm.008}$ \\
  & 25
  & $\mathbf{.692}{\scriptstyle\pm.008}$
  & $\underline{.764}{\scriptstyle\pm.002}$
  & $\underline{.727}{\scriptstyle\pm.004}$
  & $\underline{.590}{\scriptstyle\pm.022}$
  & $\underline{.707}{\scriptstyle\pm.012}$
  & $\mathbf{.581}{\scriptstyle\pm.007}$
  & $\underline{.677}{\scriptstyle\pm.005}$ \\
  & 50
  & $\underline{.700}{\scriptstyle\pm.006}$
  & $\underline{.791}{\scriptstyle\pm.000}$
  & $\underline{.748}{\scriptstyle\pm.005}$
  & $.592{\scriptstyle\pm.011}$
  & $\underline{.733}{\scriptstyle\pm.001}$
  & $\underline{.599}{\scriptstyle\pm.009}$
  & $\underline{.694}{\scriptstyle\pm.003}$ \\
  & 100
  & \underline{.712} & .814 & \underline{.772} & \underline{.629}
  & \underline{.751} & \underline{.635} & \underline{.719} \\

\addlinespace[1pt]
SleepTransformer
  & 1
  & $\underline{.591}{\scriptstyle\pm.013}$
  & $\mathbf{.685}{\scriptstyle\pm.017}$
  & $\mathbf{.624}{\scriptstyle\pm.050}$
  & $\mathbf{.550}{\scriptstyle\pm.016}$
  & $\mathbf{.596}{\scriptstyle\pm.011}$
  & $\mathbf{.511}{\scriptstyle\pm.012}$
  & $\mathbf{.593}{\scriptstyle\pm.012}$ \\
  & 10
  & $\mathbf{.671}{\scriptstyle\pm.014}$
  & $\mathbf{.769}{\scriptstyle\pm.003}$
  & $\mathbf{.713}{\scriptstyle\pm.009}$
  & $\mathbf{.578}{\scriptstyle\pm.027}$
  & $\mathbf{.689}{\scriptstyle\pm.012}$
  & $\mathbf{.549}{\scriptstyle\pm.004}$
  & $\mathbf{.661}{\scriptstyle\pm.003}$ \\
  & 25
  & $\underline{.691}{\scriptstyle\pm.007}$
  & $\mathbf{.795}{\scriptstyle\pm.002}$
  & $\mathbf{.732}{\scriptstyle\pm.011}$
  & $\mathbf{.602}{\scriptstyle\pm.018}$
  & $\mathbf{.737}{\scriptstyle\pm.008}$
  & $\underline{.574}{\scriptstyle\pm.001}$
  & $\mathbf{.689}{\scriptstyle\pm.003}$ \\
  & 50
  & $\mathbf{.706}{\scriptstyle\pm.002}$
  & $\mathbf{.813}{\scriptstyle\pm.002}$
  & $\mathbf{.754}{\scriptstyle\pm.017}$
  & $\mathbf{.620}{\scriptstyle\pm.038}$
  & $\mathbf{.778}{\scriptstyle\pm.008}$
  & $\mathbf{.609}{\scriptstyle\pm.000}$
  & $\mathbf{.714}{\scriptstyle\pm.005}$ \\
  & 100
  & \textbf{.718} & \textbf{.827} & \textbf{.783} & .608
  & \textbf{.802} & \textbf{.639} & \textbf{.730} \\

\addlinespace[1pt]
SleepFM
  & 1
  & $.535{\scriptstyle\pm.012}$
  & $.606{\scriptstyle\pm.024}$
  & $.538{\scriptstyle\pm.035}$
  & $.502{\scriptstyle\pm.014}$
  & $.524{\scriptstyle\pm.028}$
  & $.504{\scriptstyle\pm.010}$
  & $.535{\scriptstyle\pm.015}$ \\
  & 10
  & $.609{\scriptstyle\pm.007}$
  & $.704{\scriptstyle\pm.005}$
  & $.664{\scriptstyle\pm.015}$
  & $.538{\scriptstyle\pm.026}$
  & $.607{\scriptstyle\pm.021}$
  & $.541{\scriptstyle\pm.009}$
  & $.611{\scriptstyle\pm.006}$ \\
  & 25
  & $.640{\scriptstyle\pm.008}$
  & $.741{\scriptstyle\pm.004}$
  & $.711{\scriptstyle\pm.007}$
  & $.557{\scriptstyle\pm.019}$
  & $.635{\scriptstyle\pm.016}$
  & $.560{\scriptstyle\pm.004}$
  & $.640{\scriptstyle\pm.006}$ \\
  & 50
  & $.668{\scriptstyle\pm.010}$
  & $.759{\scriptstyle\pm.002}$
  & $.726{\scriptstyle\pm.000}$
  & $.571{\scriptstyle\pm.004}$
  & $.660{\scriptstyle\pm.019}$
  & $.569{\scriptstyle\pm.003}$
  & $.659{\scriptstyle\pm.001}$ \\
  & 100
  & .686 & .782 & .751 & .608 & .689 & .594 & .685 \\

\addlinespace[1pt]
OSF
  & 1
  & $.533{\scriptstyle\pm.018}$
  & $.620{\scriptstyle\pm.024}$
  & $.552{\scriptstyle\pm.026}$
  & $.509{\scriptstyle\pm.038}$
  & $.531{\scriptstyle\pm.014}$
  & $.502{\scriptstyle\pm.011}$
  & $.541{\scriptstyle\pm.011}$ \\
  & 10
  & $.608{\scriptstyle\pm.007}$
  & $\underline{.727}{\scriptstyle\pm.005}$
  & $.665{\scriptstyle\pm.008}$
  & $.548{\scriptstyle\pm.028}$
  & $.585{\scriptstyle\pm.007}$
  & $.521{\scriptstyle\pm.010}$
  & $.609{\scriptstyle\pm.006}$ \\
  & 25
  & $.645{\scriptstyle\pm.015}$
  & $.762{\scriptstyle\pm.002}$
  & $.704{\scriptstyle\pm.012}$
  & $.580{\scriptstyle\pm.056}$
  & $.603{\scriptstyle\pm.023}$
  & $.550{\scriptstyle\pm.017}$
  & $.641{\scriptstyle\pm.008}$ \\
  & 50
  & $.676{\scriptstyle\pm.001}$
  & $.786{\scriptstyle\pm.004}$
  & $.725{\scriptstyle\pm.006}$
  & $\underline{.611}{\scriptstyle\pm.039}$
  & $.663{\scriptstyle\pm.013}$
  & $.571{\scriptstyle\pm.018}$
  & $.672{\scriptstyle\pm.005}$ \\
  & 100
  & .698 & .810 & .748 & \textbf{.637} & .709 & .603 & .701 \\

\addlinespace[1pt]
Hypnos
  & 1
  & $.537{\scriptstyle\pm.018}$
  & $\underline{.642}{\scriptstyle\pm.022}$
  & $.541{\scriptstyle\pm.019}$
  & $.489{\scriptstyle\pm.022}$
  & $.532{\scriptstyle\pm.030}$
  & $.504{\scriptstyle\pm.017}$
  & $.541{\scriptstyle\pm.009}$ \\
  & 10
  & $.619{\scriptstyle\pm.019}$
  & $.726{\scriptstyle\pm.006}$
  & $.684{\scriptstyle\pm.007}$
  & $.541{\scriptstyle\pm.010}$
  & $.623{\scriptstyle\pm.018}$
  & $.526{\scriptstyle\pm.010}$
  & $.620{\scriptstyle\pm.006}$ \\
  & 25
  & $.650{\scriptstyle\pm.014}$
  & $.753{\scriptstyle\pm.008}$
  & $.712{\scriptstyle\pm.006}$
  & $.561{\scriptstyle\pm.034}$
  & $.661{\scriptstyle\pm.007}$
  & $.550{\scriptstyle\pm.009}$
  & $.648{\scriptstyle\pm.002}$ \\
  & 50
  & $.677{\scriptstyle\pm.010}$
  & $.775{\scriptstyle\pm.007}$
  & $.736{\scriptstyle\pm.002}$
  & $.585{\scriptstyle\pm.008}$
  & $.692{\scriptstyle\pm.002}$
  & $.577{\scriptstyle\pm.001}$
  & $.674{\scriptstyle\pm.000}$ \\
  & 100
  & .702 & \underline{.816} & .758 & .620
  & .715 & .597 & .701 \\

\bottomrule
\end{tabular*}
\end{table*}

\paragraph{Global-mean linear probe}\label{app:downstream_modeling}

To assess how much information remained accessible to a minimal, order-invariant linear readout, we replaced the complete downstream framework with a global-mean linear probe (Table \ref{tab:global_mean_linear}). For every representation stream, tokens were averaged across valid, non-padded time points at the encoder's native temporal resolution. Missing streams contributed zero-valued vectors and were identified by appended binary availability indicators. The stream means and availability indicators were concatenated, and a single linear layer jointly produced the six disorder logits. The probe used the same outer folds and multilabel binary-cross-entropy objective as the primary downstream models.
SleepTransformer retained the highest macro-AUROC under linear probing (.731), followed by OSF (.716), Hypnos (.701), SleepFM (.676), and U-Sleep (.613). Relative to whole-night state-space modeling, macro-AUROC decreased by .118 for U-Sleep, .046 for SleepFM, .027 for Hypnos, .013 for SleepTransformer, and .010 for OSF. Thus, richer whole-night modeling improved macro-AUROC for every encoder, but its contribution varied substantially.

\begin{table}[hbt!]
\caption{
Pooled out-of-fold AUROC obtained using the global-mean linear probe without demographic and anthropometric covariates. Best and second-best results are bolded and underlined.
}
\label{tab:global_mean_linear}
\centering
\scriptsize
\setlength{\tabcolsep}{2.0pt}
\renewcommand{\arraystretch}{0.92}

\begin{tabular*}{\linewidth}{
@{\extracolsep{\fill}}lccccccc@{}
}
\toprule
Method & Insomnia & SDB & CDH & CRSWD & Parasomnia & SRMD & Macro \\
\midrule

U-Sleep
  & .672 & .695 & .669 & .614 & .517 & .514 & .613 \\

SleepTransformer
  & \textbf{.725}
  & \underline{.845}
  & \textbf{.763}
  & \textbf{.629}
  & \textbf{.785}
  & .640
  & \textbf{.731} \\

SleepFM
  & .685 & .809 & .722 & .537 & .687 & .615 & .676 \\

OSF
  & \underline{.706}
  & \textbf{.848}
  & \underline{.750}
  & \underline{.624}
  & .717
  & \textbf{.648}
  & \underline{.716} \\

Hypnos
  & .698
  & .827
  & .738
  & .575
  & \underline{.721}
  & \underline{.646}
  & .701 \\

\bottomrule
\end{tabular*}
\end{table}


\end{document}